\pdfoutput=1

\documentclass[11pt]{article}

\usepackage{authblk}
\usepackage{lipsum}

\newcommand\blfootnote[1]{%
  \begingroup
  \renewcommand\thefootnote{}\footnote{#1}%
  \addtocounter{footnote}{-1}%
  \endgroup
}

\usepackage{ACL2023}

\usepackage{times}
\usepackage{latexsym}

\usepackage[T1]{fontenc}

\usepackage[utf8]{inputenc}

\usepackage{microtype}

\usepackage{inconsolata}

\usepackage{enumitem}
\usepackage{xcolor}
\usepackage{expex}

\title{Factoring Hate Speech: A New Annotation Framework to Study Hate Speech in Social Media}

\author[1$\dagger$]{Gal Ron}
\author[2$\dagger$]{Effi Levi}
\author[1]{Odelia Oshri}
\author[1]{Shaul R. Shenhav}
\affil[1]{Department of Political Science, The Hebrew University of Jerusalem}
\affil[2]{Institute of Computer Science, The Hebrew University of Jerusalem}
\affil[ ]{{\tt \{gal.ron2$|$odelia.oshri$|$shaul.shenhav\}@mail.huji.ac.il}}
\affil[ ]{{\tt efle@cs.huji.ac.il}}

\begin{document}
\maketitle

\blfootnote{$^\dagger$Both authors contributed equally to this work.}

\begin{abstract}
In this work we propose a novel annotation scheme which factors hate speech into five separate discursive categories. To evaluate our scheme, we construct a corpus of over 2.9M Twitter posts containing hateful expressions directed at Jews, and annotate a sample dataset of 1,050 tweets. We present a statistical analysis of the annotated dataset as well as discuss annotation examples, and conclude by discussing promising directions for future work.
\end{abstract}

\section{Introduction}
\label{sec:intro}

Social media has come to constitute a space for the propagation of hostility \cite[see][p. 1]{elsherief2018peer} and provides fertile grounds for the radicalization of individuals in support of violent extremist groups~\citep{reynolds2016counter,mitts2019isolation}. Much research has been devoted to automatically identifying hate speech in online forums~\citep{mathew2021hatexplain,kim-etal-2022-generalizable,wiegand-etal-2022-identifying} along with factors that facilitate its propagation~\citep{Scharw_chter_2020,newman2021trump}, and classifying different forms of hateful and abusive content~\citep{davidson2017automated,founta2018large}. However, the very concept of ``hate'' and its presence in written text is somewhat illusive and amorphous~\cite{fortuna-etal-2020-toxic}; therefore, some efforts have been made to define different typologies of hate speech ~\cite{waseem2017understanding,founta2018large,davidson2017automated,mathew2021hatexplain}.

In this work, we address this important subject through the specific case of hate speech directed towards Jews in Twitter posts. 
Rather than attempting to classify hate speech into different types, we present a novel annotation scheme which factors hate speech into a comprehensive set of separate discursive aspects that often appear in hate speech, capturing its intricate and diverse nature.

This factorization aims to achieve several goals: first, make the annotation process more focused and accurate, by decomposing the amorphous and ambiguous concept of ``hate'' into 
more specific and narrowly defined discourse aspects, rendering the annotation process more objective. Second, it allows exploring and analyzing hate speech across these different aspects, 
hopefully leading to a deeper understanding of its complexities, variety, and nuance. 
Furthermore, this set of aspects defines various possible distinct combinations, each of which encodes a different and unique configuration of hate speech.
Although this annotation scheme was designed to capture and characterize hate speech directed towards Jews, with the exception of one group-specific aspect, it is general enough to be applied to any other group-directed hate speech.

We constructed a corpus of Twitter conversations in English containing over 2.9M tweets, collected through Twitter API v2. 
In order to evaluate our annotation scheme on real Twitter posts, we used it to annotate a sample of 1,050 tweets taken from the corpus. We present a quantitative analysis of the annotated dataset, as well as a qualitative one (through the use of some examples). We conclude by discussing several directions to extend and develop our work.

{\color{red} \textbf{Content Warning}: This document contains some examples of hateful content. This is strictly for the purpose of enabling this research. 
Please be aware that this content could be offensive and cause you distress.}

\begin{table*}[!htbp]
\centering
\begin{tabular}{l|c|cc|c|cc}
    & \textbf{Single Tweets} & \multicolumn{2}{c|}{\textbf{Conversations}} & \textbf{Total} & \multicolumn{2}{c}{\textbf{Conversation Length}} \\
    \cline{2-7}
    & \textbf{\# Tweets} & \textbf{\# Conversations} & \textbf{\# Tweets} & \textbf{\# Tweets} & \textbf{Mean} & \textbf{STD} \\
    \hline 
    \textbf{Neutral} & 601,917 & 109,172 & 1,005,095 & 1,607,012 & 9.21 & 71.73 \\
    \textbf{Racial}  & 527,541 & 97,730 & 788,576 & 1,316,117 & 8.07 & 127.43 \\
    \hline
    \textbf{Total} & 1,129,458 & 206,902 & 1,793,671 & 2,923,129 & & \\
\end{tabular}
\caption{Complete tweet corpus statistics. ``Single Tweets'': tweets that were posted as new tweets rather than as a reply, and were not replied to.}
\label{tab:corpus_stats}
\end{table*}

\section{Tweets Corpus}
\label{sec:corpus}

The tweets were extracted through the Twitter API v2 using the \texttt{tweepy} python module~\footnote{\url{https://github.com/tweepy/tweepy}}. We applied for, and were granted, an Academic Research Access to the API~\footnote{No longer available, as of May 2023}, which offers a full-archive access to public data posted on the platform, going back to April 2006.

To increase the likelihood of retrieving tweets that contain expressions of hate towards Jews, we used two types of keyword-based filters in our queries: \textit{neutral} keywords and \textit{racial} keywords. The neutral stop list -- containing 14 words -- was compiled from keywords referenced in previous studies~\citep{gunther2021detecting,chandra2021subverting}. The racial stop list -- containing 28 words and expressions -- was compiled using the \texttt{Hatebase} database, a multilingual lexicon for racial terms~\footnote{\url{https://hatebase.org}}, by extracting from the database all the English terms pertaining to Jews and Judaism that had at least one sighting.

Following preliminary experimentation with the API directed at increasing the chances of retrieving a conversation (thread) containing Jew-related hate expressions, we decided to focus on collecting conversations which stemmed from a ``source'' tweet (a new post rather than a reply to another tweet) adhering to our keyword filters.
We devised the following 2-step process. Given a specific date (24-hour interval) and a specific keyword filter:

\begin{enumerate}
    \item Query the API for English ``source'' tweets containing any of the keywords in the filter, posted within the specified date.
    \item For every tweet extracted in step 1: if the tweet was replied to, query the API for all the available tweets in the resulting conversation (some tweets, such as deleted or private tweets, were not available for extraction).
\end{enumerate}

For each date between July 1\textsuperscript{st} 2018 and June 30\textsuperscript{th} 2022 (defining a period of exactly 4 years), we applied the procedure with the \textit{racial} keywords filter and collected as many tweets as possible. Then, we applied the same procedure with the \textit{neutral} keyword filter to collect a similar number of tweets from the same date~\footnote{preliminary experiments showed that tweets containing the \textit{neutral} keywords are significantly more abundant compared to the \textit{racial} keywords}. This was done in order to keep the corpus as balanced as possible between the two types of keyword filters. 

The result is a large corpus of Twitter conversations started between July 1\textsuperscript{st} 2018 and June 30\textsuperscript{th} 2022, segmented by the type of the filter applied to the conversation's source tweet (\textit{neutral} or \textit{racial} Jews-related keywords). 
Aside from the text itself, the corpus includes additional meta-data for each tweet: tweet ID, conversation ID, posting date, reply-to ID (if the tweet was written as a reply to another tweet), tweet statistics (retweets, replies, likes, quotes and views), place \& country (if available), author ID (a unique identifier for the author of the tweet) and author statistics (followers, verified status).
Note that the ``conversation ID'' and ``reply-to ID'' fields allow a complete hierarchical reconstruction of a conversation given any tweet from that conversation.

Statistics for the corpus are given in Table~\ref{tab:corpus_stats}.

\section{Hate Speech Annotation}

\begin{table*}[!htbp]
\centering
\begin{tabular}{l|c|cc|c}
    & \textbf{Single Tweets} & \multicolumn{2}{c|}{\textbf{Conversations}} & \textbf{Total} \\
    \cline{2-5}
    & \textbf{\# Tweets} & \textbf{\# Conversations} & \textbf{\# Tweets} & \textbf{\# Tweets} \\
    \hline 
    \textbf{Neutral} & 263 & 82 & 263 & 526 \\
    \textbf{Racial}  & 262 & 82 & 262 & 524 \\
    \hline
    \textbf{Total} & 525 & 164 & 525 & 1,050 \\
\end{tabular}
\caption{Statistics for the sample dataset}
\label{tab:dataset_stats}
\end{table*}

\subsection{Annotation Scheme}
\label{subsec:annotation_scheme}
As discussed in Section~\ref{sec:intro}, we have devised a novel annotation scheme with the goal of factoring hate speech into several separate aspects.
The scheme encodes five different discursive categories, which are designed to capture the main recurring aspects of hate speech as employed and defined in previous studies~\citep{kaid2007encyclopedia,davidson2017automated,arango2022multilingual,khurana-etal-2022-hate}, as well as the discursive elements of hate speech that are described in the United Nations' "Strategy and Plan of Action on Hate Speech"~\footnote{\url{https://www.un.org/en/hate-speech/understanding-hate-speech/what-is-hate-speech}}. An aggregative definition, as suggested here, enables us to identify hate speech towards the target group in a broad yet nuanced sense, while also differentiating between various forms of expression. 
Importantly, our annotation scheme aims to capture clear expressions of hate rather than mere hateful terms. As such, tweets containing the racial keywords which were used in our queries to the Twitter API were annotated under the relevant category only when it was plausible to suspect that these words were indeed employed to express hate.

\begin{table}[!htbp]
\centering
\begin{tabular}{lc}
    \hline 
    \textbf{Category} & \textbf{\# Tweets} \\
    \hline
    Contempt & 5 \\
    Abuse & 181 \\
    Call for Anti-Group Action & 31 \\
    Prejudice & 12 \\
    Holocaust Denial & 4 \\
    \hline 
\end{tabular}
\caption{Annotation statistics for the sample dataset}
\label{tab:annotation_stats}
\end{table}

The five categories are:

\begin{enumerate}
    \item \textbf{Contempt} -- speech that conveys a strong disliking of, or negative attitudes towards the targeted group, and does so in a neutral tone or form of expression.
    \item \textbf{Abuse} -- speech that demeans, degrades, vilifies, mocks, humiliates, or conveys general hostility that is expressed using emotionally-charged language.
    \item \textbf{Call for Anti-Group Action} -- an incitement of violence and/or discrimination against the target group.
    \item \textbf{Prejudice} -- the expression of negative thoughts/beliefs regarding the targeted group on the basis of the group’s characteristics, and/or (negative) monolithic references to the targeted group.
    \item \textbf{Holocaust Denial} -- the only category specific to our target group (Jews), this includes derecognition of the holocaust, or statements that recognize the fact that the holocaust happened but degrade from its scope, mock it (and/or the people it hurt), and belittle its significance. 
\end{enumerate}

These discursive categories not only encompass substantive elements of hate, such as contempt and prejudice, but also address the manner in which negative discourse is conveyed, including abusive language and incitement of violence.

All the categories, except the fifth (\texttt{Holocaust Denial}), are general and may naturally be applied to other groups besides our target group (Jews). In addition, while the categories encode separate aspects of hateful discourse, they may conjointly characterize the same expression; for example, a post can be abusive while also expressing prejudice. Consequently, the annotation scheme defines a \textit{multi-label} classification task. 

\subsection{Annotated Sample Dataset}

\begin{table*}[!htbp]
\centering
\begin{tabular}{lcccc}
    \hline
    & & & \textbf{Call for Anti-} & \\
    & \textbf{Contempt} & \textbf{Abuse} & \textbf{Gruop Action} & \textbf{Prejudice} \\
    \hline 
    \textbf{Abuse} & -0.0316 & & & \\
    \textbf{Call for Anti-Gruop Action} & -0.0121 & 0.2332 & & \\ 
    \textbf{Prejudice} & 0.1227 & 0.1644 & 0.0342 & \\
    \textbf{Holocaust Denial} & -0.0043 & 0.0536 & -0.0108 & 0.1388 \\
    \hline
\end{tabular}
\caption{Inter-category Pearson's correlations in the sample dataset}
\label{tab:annotation_corrs}
\end{table*}

For the purpose of conducting a preliminary analysis of our annotation scheme over real tweet data, we annotated a sample dataset of 1,050 tweets from the corpus described in Section~\ref{sec:corpus}. These tweets were sampled by iterating the dates backwards, starting from June 30\textsuperscript{th} 2022. For each date, one conversation with a length of $2 \leq k \leq 10$ tweets was randomly selected, then $k$ additional single tweets (1-tweet conversations) were randomly selected from the same date. This procedure was performed separately for each of the two keyword filter types (\textit{neutral} and \textit{racial}).  The result is a collection of 1,050 tweets from conversations started between June 4\textsuperscript{th} 2022 and June 30\textsuperscript{th} 2022 (statistics are given in Table~\ref{tab:dataset_stats}). Each tweet was encoded with a subset of the five possible categories (described in Section~\ref{subsec:annotation_scheme}), including the empty set (none of the categories). 

Table~\ref{tab:annotation_stats} shows the annotation statistics for the sample dataset. Note that despite the use of keyword filters to retrieve the tweets from the Twitter API, all five categories are generally sparse in the dataset. The most common category is \texttt{Abuse}, with 181 instances (out of the 1,050 tweets in the dataset). 
Given that half of these tweets were collected using the \textit{racial} keywords filter, this is consistent with previous findings that hate speech is highly likely to contain racial slurs~\citep{davidson2017automated}; intuitively, it is the most ``direct'' way to express hate (among the five categories). It is important to note, however, that the mere presence of a racial slur does not automatically merit an \texttt{Abuse} annotation, as evident in the following example:

\ex
In the span of 5 min I have been both called a "toxic fan" for not liking the Kenobi show and a "Zionazi" by a fan of the show. Maybe you should reconsider who are the "toxic" ones (\texttt{None})
\xe

The second most common category is \texttt{Call for Anti-Group Action}, followed by \texttt{Prejudice} and \texttt{Contempt}, with the most uncommon category being \texttt{Holocaust Denial}.

Table~\ref{tab:annotation_corrs} displays the inter-category correlations (measured in Pearson's \textit{r}) in the sample dataset. In general, no considerable correlations were found between any two categories. \texttt{Abuse} was found to be somewhat correlated with \texttt{Call for Anti-Group Action} ($r=0.2332$), and to a lesser degree with \texttt{Prejudice} ($r=0.1644$). Naturally, calls for violence and prejudiced expressions are often accompanied by abusive language, for example:

\ex
Ahhhhhh the good old days , yids were bums then still bums now. (\texttt{Abuse}, \texttt{Prejudice})
\xe

A minor correlation between \texttt{Prejudice} and \texttt{Contempt} ($r=0.1227$) is possibly an indication that prejudiced perspectives serve as a kind of "rationale" for hate that does not always require the more emotional use of abusive language. This is demonstrated in the following example:

\ex
Perhaps America is just too fat, spoiled and lazy not to be noosed by the Jews into another low road oblivion that profits the jews. It seems incapable of tweaking itself or nurturing and governing through its' higher itself. Better perhaps to help it rot? (\texttt{Prejudice}, \texttt{Contempt})
\xe

Another minor correlation between \texttt{Prejudice} and \texttt{Holocaust Denial} ($r=0.1388$) may be attributed to the fact that both categories are closely associated with conspiracy theories.

\section{Conclusion}

In this paper, we address the task of capturing and characterizing hate speech directed towards Jews in Twitter posts. For that purpose, we devised a novel annotation scheme that encodes five different aspects of hate speech, four of which are not specific to our target group (Jews), allowing us to factorize the generally amorphous concept of ``hate'' into more concretely defined aspects. We utilized the Twitter API v2 to collect and assemble a corpus of Twitter conversations in English containing over 2.9M tweets, using two types of keyword filters (\textit{neutral} and \textit{racial}) to maximize the likelihood of retrieving tweets that contain expressions of hate towards Jews. We then used our annotation scheme to annotate a sample of 1,050 tweets, and demonstrated its potential 
contribution through select examples. We intend to make all of these resources (tweet corpus, annotation guidelines and sample dataset) available to the research community.

We are currently engaged in an ongoing effort to train additional annotators and use our assembled tweet corpus to produce a large and comprehensive annotated dataset. We are also working -- in parallel -- on assembling and annotating a similar corpus for Muslim-related hateful expressions.

Another direction we are currently pursuing is taking advantage of the fact that the corpus is comprised of complete Twitter conversations, to annotate expressions of hate in the context of the conversation which the tweet is a part of (rather than just based on the content of the tweet itself). For example, replying to a hateful post with strong agreement may be considered as hate speech only if the context (preceding posts) is taken into account. Using the hierarchical structure of the conversation will allow not only encoding such cases, but also modelling the dynamics of hate speech as it progresses through the conversation and over time.

In addition, we plan to utilize the tweet corpus to explore \textit{counter-hate} speech~\citep{benesch2016considerations,wright2017vectors,garland-etal-2020-countering}. As these types of expressions are reactive by nature, complete Twitter conversations are instrumental in addressing and analyzing them. Augmenting hate-speech annotated Twitter conversations with counter-hate annotation will allow us to explore the inter-changing dynamics of hate and counter-hate speech,
as well as which kinds of counter messages are tailored to the different hate categories. We might find for example that the effective counter messages for abusive speech are those that attack the user, while most effective counter messages for prejudice deliver data and facts to contradict the prejudiced beliefs.

\section*{Limitations}

There are three main limitations to our work.
One limitation results directly from the single target group included in our analysis (Jews). While our annotation scheme was designed to be as general as possible (with the exception of the \texttt{Holocaust Denial} category), applying it to a single target group does not allow us to evaluate the extent of its generalizability to other target groups. 

A second limitation has to do with messages that support and fuel hate, without containing actual hateful content (expressing agreement with another hateful message).
While such messages may spread hate, they would not be encoded in our annotation setup, since  the message context (e.g., the surrounding conversation) is not taken into account during the annotation process.

Thirdly, our annotation scheme does not currently account for how the annotated hate is perceived by the message's readers. This information may lie in the reaction incurred by the hateful message -- the tweet's replies, as well as the its meta-data (number of likes, quotes, etc.).

By applying the annotation to other target groups, and by annotating complete conversations (thus capturing the context of the tweets), 
future research could address the three limitations.

\section{Acknowledgments}

We wish to thank Prof. Peter J. Loewen, Prof. Ron Levi, Prof. Christopher Cochrane, and Mr. Thomas Bergeron for their valuable contribution. 

This research was partially funded by the University of Toronto - Hebrew University of Jerusalem Research and Training Alliance, and by the Knapp Family Foundation Doctoral Fellowship at the Vidal Sassoon International Center for the Study of Antisemitism.

\bibliography{acl2023}
\bibliographystyle{acl_natbib}

\end{document}